\documentclass[runningheads]{llncs}

\usepackage{eccv}

\usepackage{eccvabbrv}

\usepackage{graphicx}
\usepackage{booktabs}

\usepackage[accsupp]{axessibility}  % Improves PDF readability for those with disabilities.

\usepackage{float}
\usepackage{amsmath}
\DeclareMathOperator*{\argmin}{arg\,min}
\usepackage[utf8]{inputenc}
\usepackage{booktabs}
\usepackage{bm}

\usepackage{graphicx}

\usepackage[pagebackref,breaklinks,colorlinks,citecolor=eccvblue]{hyperref}
\usepackage{orcidlink}

\newcommand\NAME{\texttt{UrbanDiffusion}\xspace}

\begin{document}

\title{Urban Scene Diffusion through Semantic Occupancy Map} 


\author{
Junge Zhang$^{1,4}$ \thanks{} \;
Qihang Zhang$^2$ \;
Li Zhang$^1$ \;
Ramana Rao Kompella$^3$ \;
Gaowen Liu$^3$ \;
Bolei Zhou$^4$
% \\
% $^1$Fudan University \
% $^2$The Chinese University of Hong Kong \
% $^3$Cisco \\
% $^4$University of California, Los Angeles
}
\footnotetext[1]{The work was done when Junge Zhang was a visiting graduate student at UCLA.}

\authorrunning{Zhang et al.}

% % TODO FINAL: Replace with your institution list.
\institute{$^1$Fudan University \
$^2$The Chinese University of Hong Kong \
$^3$Cisco \\
$^4$University of California, Los Angeles
}

\maketitle

\begin{abstract}

Generating unbounded 3D scenes is crucial for large-scale scene understanding and simulation. Urban scenes, unlike natural landscapes, consist of various complex man-made objects and structures such as roads, traffic signs, vehicles, and buildings. 
To create a realistic and detailed urban scene, it is crucial to accurately represent the geometry and semantics of the underlying objects, going beyond their visual appearance. 
In this work, we propose \NAME, a 3D diffusion model that is conditioned on a Bird’s-Eye View (BEV) map and generates an urban scene with geometry and semantics in the form of semantic occupancy map. Our model introduces a novel paradigm that learns the data distribution of scene-level structures within a latent space and further enables the expansion of the synthesized scene into an arbitrary scale. After training on real-world driving datasets, our model can generate a wide range of diverse urban scenes given the BEV maps from the held-out set and also generalize to the synthesized maps from a driving simulator. We further demonstrate its application to scene image synthesis with a pretrained image generator as a prior. The project website is \url{https://metadriverse.github.io/urbandiff/}.
\end{abstract}    
\section{Introduction}
\label{sec:intro}

Significant progress has been made in image generation and 3D-aware object generation~\cite{stablediff,dalle,eg3d,giraffe,dreamfusion}. Most of the works focus on generating individual objects or scenes in a single image. However, large-scale urban scene generation, with its application to interactive scene simulation and autonomous driving, remains much less explored. Different from natural scene generation~\cite{scenedreamer,infinitenature,persistentnature}, generating urban scenes requires composing accurate geometry and semantics of the man-made objects and structures such as roads, lanes, buildings, and traffic signs. Many recent attempts have achieved success in generating realistic visual appearances of urban scenes at the pixel level, including BEVGen\cite{bevgen}, BEVControl~\cite{bevcontrol} and MagicDrive~\cite{magicdrive}, as well as those video generation methods such as StyleSV~\cite{stylesv}, GAIA-1~\cite{gaia} and Drivedreamer~\cite{drivedreamer}. 

Despite the visually appealing results produced by the aforementioned methods, a fundamental limitation remains - the lack of accurate geometry and semantic information for the generated scene structure. As a result, it becomes challenging to manipulate the camera pose and viewpoint of the generated images, as well as transform the visual scene into an interactive simulation environment. Recent works like DiscoScene ~\cite{disco} and UrbanGIRAFFE~\cite{urbangiraffe} have achieved controllable 3D-aware image generation via manipulating the given scene layout, revealing the importance of grounding image generation in the underlying 3D structure for scene generation tasks.

\begin{figure}[t]
    \centering
\includegraphics[width=1\linewidth]{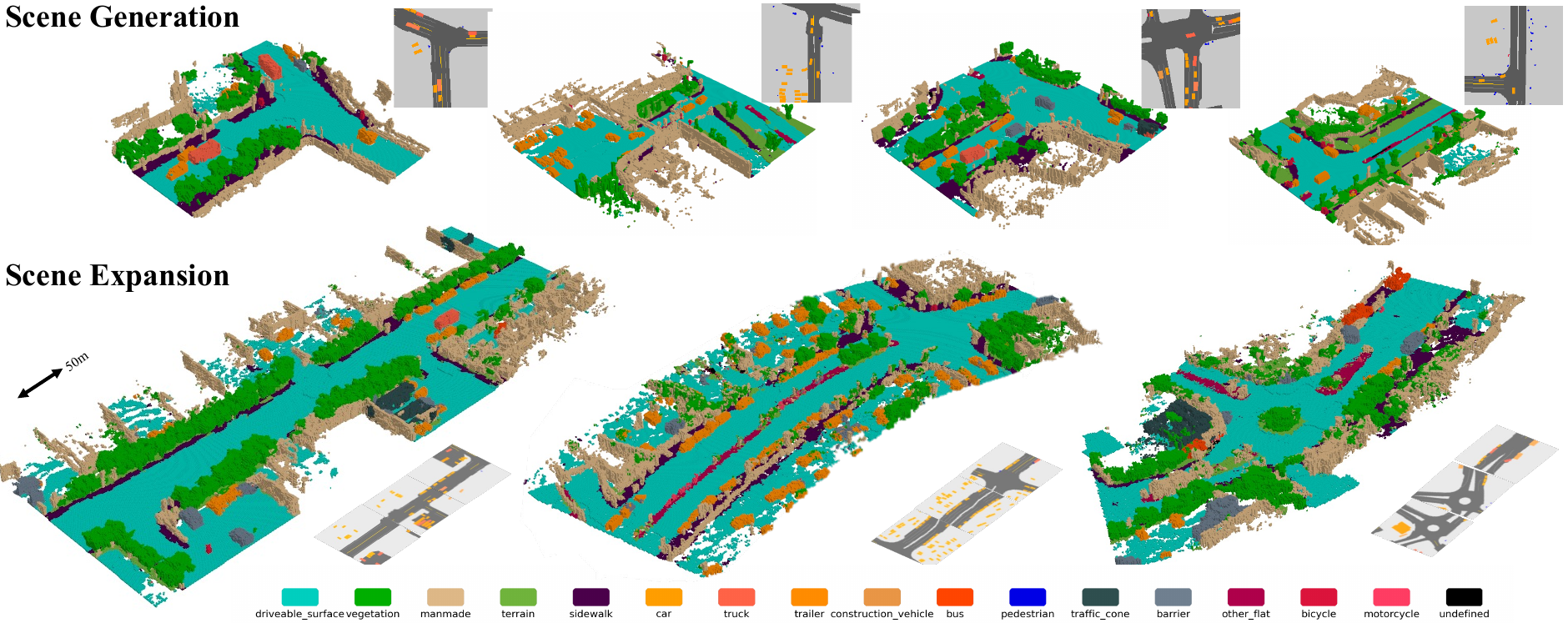}
  \captionsetup{type=figure,font=small}
  \vspace{-2mm}
  \caption{Diverse individual scenes and large-scale scenes generated by \NAME. A scene is represented by the semantic occupancy map and the color labels indicate different semantic categories. The input BEV layout is also attached as a reference.} 
  \label{fig:teaser}
  \vspace{-10mm}
\end{figure}

Since a 3D scene contains a much richer structure than a 2D image snapshot, representing 3D scenes beyond the pixels is essential for large-scale scene generation. In recent two years, the semantic occupancy map has become an effective scene representation that incorporates both geometric structure and semantics, leading to much progress in the 3D perception of autonomous driving~\cite{voxformer,occ3d,surroundocc}. Occupancy map as scene representation allows the driving system to extract necessary scene information for planning and navigation~\cite{scene_as_occ,stp3}. Therefore, generating 3D scenes in terms of occupancy maps can not only incorporate geometric structure and semantics but also facilitate the future integration of the generated scenes for downstream tasks such as interactive environment simulation and scene image rendering~\cite{urbangiraffe}.

In this work, we propose \NAME that considers occupancy maps as the scene representation and learns to generate large-scale urban scenes through a 3D diffusion model. Fig.~\ref{fig:teaser} shows diverse single-frame scenes and large-scale scenes generated by the proposed method. \NAME takes the Birds' Eye View (BEV) layout as the input condition, which describes a traffic scenario, and then generates the corresponding 3D semantic occupancy maps as the output.

We first embed the collected semantic occupancy map from the real-world datasets and learn a denoiser in the latent space through the diffusion process. We then design an extensible representation to aggregate multiple frames for large-scale scene generation. The experiments show that the proposed model can generate diverse and realistic urban scenes from complex BEV layouts sampled from the validation set of the NuScenes dataset~\cite{nuscenes} and generalize to the BEV layouts synthesized by a driving simulator MetaDrive~\cite{metadrive}. We further demonstrate that the trained model can work as an effective generative prior for downstream tasks like scene completion, scene out-painting, and scene image rendering. 

We summarize our \textbf{contributions} as follows:
\textbf{(i)}
We introduce a new task of generating unbounded urban scenes at the occupancy map level to preserve the scene geometry and semantic information. 
\textbf{(ii)}
We propose a novel 3D diffusion model conditioned on BEV map, which can generate large-scale 3D urban scenes through semantic voxels with temporal consistency.
\textbf{(iii)}
After training the proposed model can generate diverse and realistic urban scenes based on the input BEV condition and benefits downstream generation task such as point cloud segmentation and scene image synthesis.

\section{Related Work}

\noindent{\textbf{Unbounded Scene Generation}} Many image generation works have explored generating 2D images with infinite resolution~\cite{infinitygan,alis}. Other works have aimed to synthesize natural scenes with temporal consistency, including camera-fly trajectories, by applying geometric constraints during the generation of large-scale natural scenes~\cite{infinitenature,infinitenaturezero,diffdreamer,gfvs,scenedreamer,persistentnature}. However, urban scenes are very different from natural scenes because there are many man-made objects like roads, lanes, buildings, and traffic signs, which are in rigid structure and arrangement. Generating realistic urban scenes requires accurate geometry and semantic information for the underlying scene structure. 
Another line of works explores generating 3D scenes from a scene layout. Notable works in this domain, such as GANCraft~\cite{gancraft} and UrbanGIRAFFE~\cite{urbangiraffe}, have demonstrated the capability to manipulate camera poses for rendering novel view images. However, it involves the prerequisite of a 3D layout, a condition that can be challenging to fulfill in real-world scenes. Recent works Infinicity~\cite{infinicity} and SGAM~\cite{SGAM} create 3D scenes by generating satellite images and then mapping them into 3D worlds. 
BerfScene~\cite{berfscene} proposes equivarient scene representation that conditions on BEV maps for unbounded scene generation.
While these previous studies have offered valuable perspectives on utilizing satellite images to reconstruct scene structures, the task of accurately recovering intricate and slender structures, such as poles, traffic signs, and other man-made features at street-level, solely from a satellite view, remains a formidable challenge. Furthermore, the ability to easily control the generated results to fit the requirements of various simulations is a critical need for the generation process. Our proposed method allows user to generate and expand urban scenes using the easily accessible Bird's Eye View (BEV) maps as input condition. 

\noindent{\textbf{3D Generation}}
Generating 3D data has been a hot topic in recent years ~\cite{pvd,dpm,chou_diffusionSDF,sdffusion,li_diffusionsdf,LION}. Some works try to learn 3D data distribution directly. PVD~\cite{pvd}, DPM~\cite{dpm}, and LION~\cite{LION} learn point diffusion, while SDF fusion works~\cite{chou_diffusionSDF,sdffusion,sdffusion} learn diffusion in SDF space. Other works like Dreamfusion~\cite{dreamfusion} and Magic3D~\cite{magic3d} recover potential 3D structures from image space with the prior of the generative model. However, these prior works are restricted to object-level generation. Some other works have attempted to perform scene-level generation, including indoor scene generation ~\cite{text2room,roomdreamer,gaudi,scenewiz3d} and outdoor scene generation~\cite{neuralfieldldm}. The most related work, NeuralField-LDM~\cite{neuralfieldldm}, has attempted to generate 3D feature volume from image space through a diffusion process. Nevertheless, as demonstrated in their paper, the complexity and ambiguity of feature volume prevent the method from depicting good geometry. The generation is also constrained within a defined range, a limitation particularly evident when attempting to generate large-scale scenes.  

\section{Method}

\begin{figure}[ht!]
    \centering
    \vspace{-10mm}
    \includegraphics[width=1\linewidth]{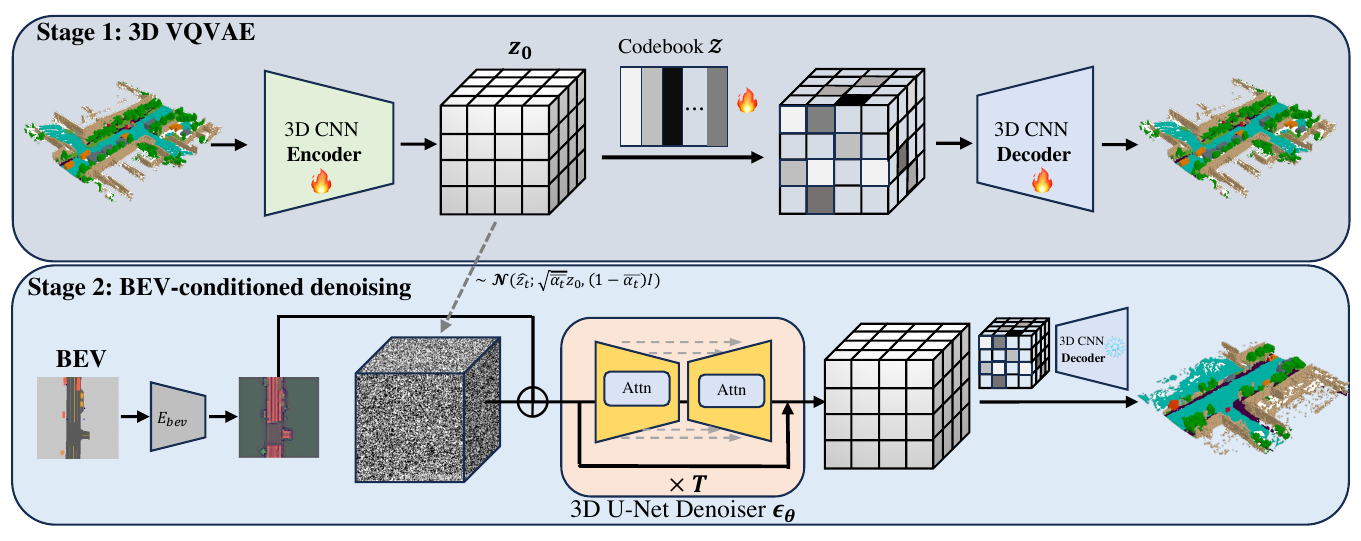}
    \caption{
    \textbf{Framework of \NAME.} An autoencoder with 3D VQVAE architecture is trained to embed semantic occupancy maps into a latent space (top). A random latent code is gradually diffused by a BEV-conditioned denoising procedure and then decoded into a semantic occupancy map (bottom).}
    \label{fig:system_fig}
    \vspace{-5mm}
\end{figure}

\NAME consists of two key components for large-scale urban scene generation: (1) BEV-conditioned 3D semantic occupancy map generation, and (2) scene extension for unbounded generation. Sec.~\ref{sec:preliminary} briefly introduces the diffusion process and guidance. Sec.~\ref{sec:generation} then presents the design of a diffusion model that incorporates the BEV map as a conditioning feature and generates 3D semantic occupancy maps. Sec.~\ref{sec:extension} finally gives the detail of the scene extension, where we leverage the capability of the trained model to expand a single local 3D scene to a large-scale scene. 

\subsection{Preliminary}
\label{sec:preliminary}
\noindent \textbf{Diffusion process}
Diffusion process~\cite{ddpm,song2020score}, which is defined as a Markov chain, can be divided into a forward process and a reverse process. In the forward process, the goal is to gradually add noise to the input data $x_0$ such that it becomes a random sample from a Gaussian distribution. The probability distribution of the forward process is determined by $\alpha_t$, which is a series of parameters to ensure the sequence converges to the Gaussian distribution: 
\begin{equation}
     \mathbf{\hat{x}}_t  = \sqrt{\alpha_t}\mathbf{\hat{x}}_{t-1} + \sqrt{1 - \alpha_t}\bm{\epsilon}_{t-1}^* \sim \mathcal{N}(\mathbf{\hat{x}}_{t} ; \sqrt{\bar\alpha_t}\mathbf{x}_0, \left(1 - \bar\alpha_t\right)\mathbf{I}), 
\end{equation}
when $\bar\alpha_t = \prod_{s=1}^t \alpha_s$. The optimization object is to predict the noise $\boldsymbol{\epsilon}$ applied to the input $x_0$ with the denoising model $\boldsymbol{\epsilon}_\theta$, $i.e.$,
\begin{equation}
    \mathbb{E}_{\boldsymbol{\epsilon},t} \|\boldsymbol{\epsilon} - \boldsymbol{\epsilon}_\theta(\mathbf{x}_t,t)\|,
\end{equation}
in a uniform sampled interval $t\in[0,1]$.

\noindent \textbf{Classifier-free Guidance}
The classifier-free guidance~\cite{classifierfree} allows the model to generate samples from a conditional distribution with high diversity. Given condition $\mathbf{c}$, the conditional probability distribution can be denoted as:
\[\tilde{\boldsymbol{\epsilon}}_\theta\left(\mathbf{x}_t, \mathbf{c}\right)=(1+w) \boldsymbol{\epsilon}_\theta\left(\mathbf{x}_t, \mathbf{c}\right)-w \boldsymbol{\epsilon}_\theta\left(\mathbf{x}_t\right).\]

\subsection{Latent Diffusion for Semantic Occupancy Map}
\label{sec:generation}

We utilize 3D semantic data represented as $x\in \{0,1\}^{H\times W\times Z\times C}$, where $H$, $W$, and $Z$ denote the length, width, and height of the voxel grids, respectively, and $C$ represents the number of semantic label categories. Inspired by the latent diffusion model (LDM)~\cite{stablediff}, we aim to train the diffusion model in a fast and memory-efficient manner, considering the high memory cost associated with representing 3D data. To achieve this, we embed the 3D semantic data $x$ into a lower-dimensional latent space, reducing memory usage and computational requirements. The LDM then utilizes this latent representation to conduct the diffusion process with a classifier-free guidance.

\noindent{\textbf{Latent representation}}
To embed the 3D semantic data into a lower-dimensional space, we adopt VQVAE~\cite{vqvae} with 3D convolution operators to regularize the real-world data that are collected from the noisy sensors. The goal is to embed the data into a latent space while maintaining the data geometry and semantics and improving the efficiency of diffusion.
We denote the operator of the encoder, the vector quantization, and the decoder respectively as $E$, $Q$, and $D$. $\mathcal{Z}$ is the corresponding codebook for the vector quantization. The training loss of VQVAE is:
\begin{equation}
   \begin{aligned}        \mathcal{L}_{\mathrm{VQVAE}}=\|\mathbf{x}-\tilde{\mathbf{x}}\|+\left\|\operatorname{sg}[E(\mathbf{x})]-\mathbf{z}_q\right\|_2^2 +\beta\left\|\operatorname{sg}\left[\mathbf{z}_q\right]-E(\mathbf{x})\right\|_2^2,  
   \end{aligned}
\end{equation}
where $\operatorname{sg}[\cdot]$ denotes stopping gradient operation in VQVAE and 

\begin{equation}
    \mathbf{z}_q = Q(E(\mathbf{x})) := \argmin\limits_{\mathbf{z}_k \in \mathcal{Z}} \|\mathbf{z}_k - E(\mathbf{x})\|, \\
     \tilde{\mathbf{x}} = D(Q(E(\mathbf{x}))),
     \label{eq:vqvae}
\end{equation}
where $\mathbf{z}_q$ denotes the latent from the codebook closest to the embedded vector $E(\mathbf{z})$ and $\tilde{\mathbf{x}}$ means the reconstructed vector.
To better align the attribute of semantic logits space, we take cross entropy as the reconstruction loss between $\mathbf{x}$ and $\tilde{\mathbf{x}}$. The quantization loss regularizes the potential noise of the data, benefiting the following diffusion process.  Later we conduct an ablation study for a proper choice of the hyper-parameters for the autoencoder in Tab.~\ref{tab:ablation_ae}.
 
\noindent{\textbf{Diffusion process}}
To accommodate the complexity of the 3D data and capture the spatial feature of the 3D structure, we adopt a standard 3D U-Net $\boldsymbol{\epsilon}_\theta$ to predict the noise for our 3D diffusion model. For the given semantic occupancy map $\mathbf{x}_0$, we denote the learning process of the diffusion model as:
\begin{equation}
    \mathbb{E}_{\boldsymbol{\epsilon}, t}\left[\left\|\hat{\boldsymbol{\epsilon}}_\theta\left(\mathbf{z}_t\right)-\boldsymbol{\epsilon}\right\|_2^2\right], \ \mathbf{z}_t = \sqrt{\bar\alpha_t}E(\mathbf{x}_0) + \left(1 - \bar\alpha_t\right)\mathbf{I}.
\end{equation}

\begin{figure}
    \centering
    \vspace{-3mm}
    \includegraphics[width=1\linewidth]{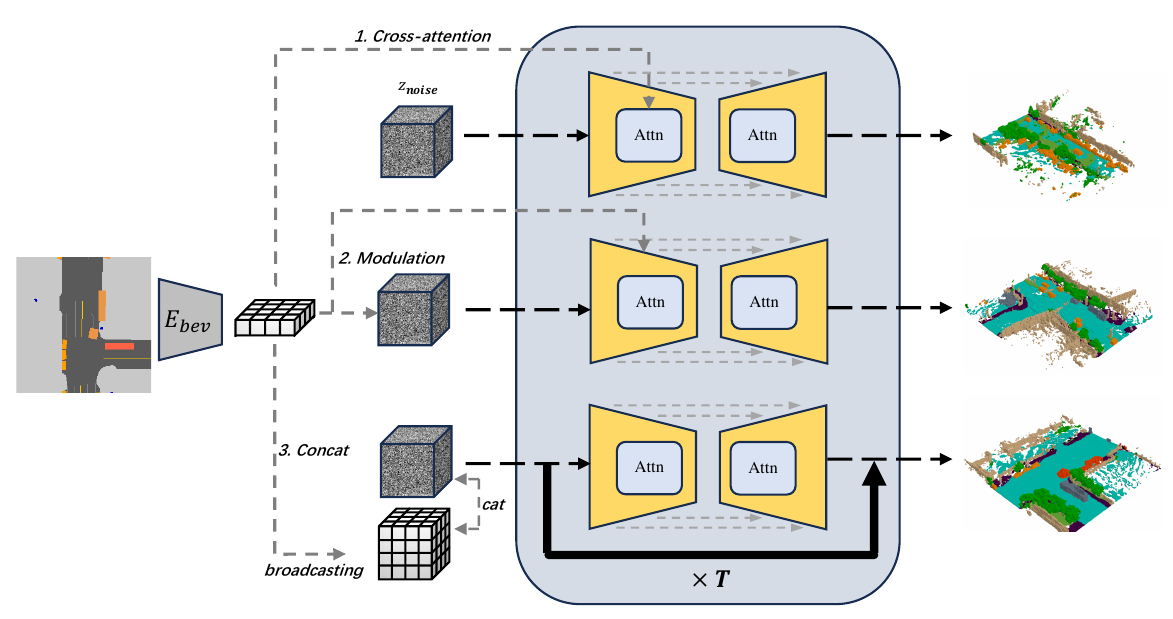}
    \caption{Different ways of BEV condition injection.}
    \label{fig:bev_inject}
    \vspace{-5mm}
\end{figure}

\noindent{\textbf{Incorporating condition into the diffusion process}}
BEV layout, which describes the traffic activities and surroundings, is crucial to the urban scene generation. It also allows the user to control 3D scene generation. Taking into account the easy accessibility of BEV maps from current simulator like MetaDrive~\cite{metadrive} and recognizing their significance in generating realistic urban environments, we propose the incorporation of BEV maps as a guiding condition in our generation process, enhancing the quality and practicality of the generated content. To figure out how to effectively inject the conditional information of BEV maps, we explore several different conditioning strategies: Cross-attention, Modulation, and Concatenate. With the condition embedding from a 2D BEV map $b \in \mathbb{R}^{H\times W\times C}$, $c_{bev} = E_{bev}(b) \in \mathbb{R}^{h\times w\times c}$,
the results of different ways of injecting conditions in Fig.~\ref{fig:bev_inject} into the generation model are shown in Tab.~\ref{tab:ablation_cond}. We find that concatenating the 2D layout BEV feature leads to better control for the large-scale scene generation shown in Fig.~\ref{fig:bev_cond}, which implies better feature alignment.

\noindent\textbf{BEV-conditional scene generation}
Given a BEV layout, the trained model can generate diverse and realistic samples that contain the scene geometry and semantic information. We adopt classifier-free guidance~\cite{classifierfree} to better conduct the conditional sampling. The noise prediction could be formulated as:
\begin{equation}
    \hat{\boldsymbol{\epsilon}}_\theta ( \mathbf{z}_t |c_{bev})=  \boldsymbol{\epsilon}_{\theta}(\mathbf{z}_t|\phi) + w \cdot (\boldsymbol{\epsilon}_{\theta}(\mathbf{z}_t|c_{bev})-\boldsymbol{\epsilon}_{\theta}(\mathbf{z}_t|\phi)).
\end{equation}
$\boldsymbol{\epsilon}_{\theta}(\mathbf{z}_t|\phi)$ means the prediction of unconditional generation.

\subsection{Scene Extension Module}
\label{sec:extension}

\begin{figure}[ht!]
    \centering
    \vspace{-2mm}
    \includegraphics[width=1\linewidth]{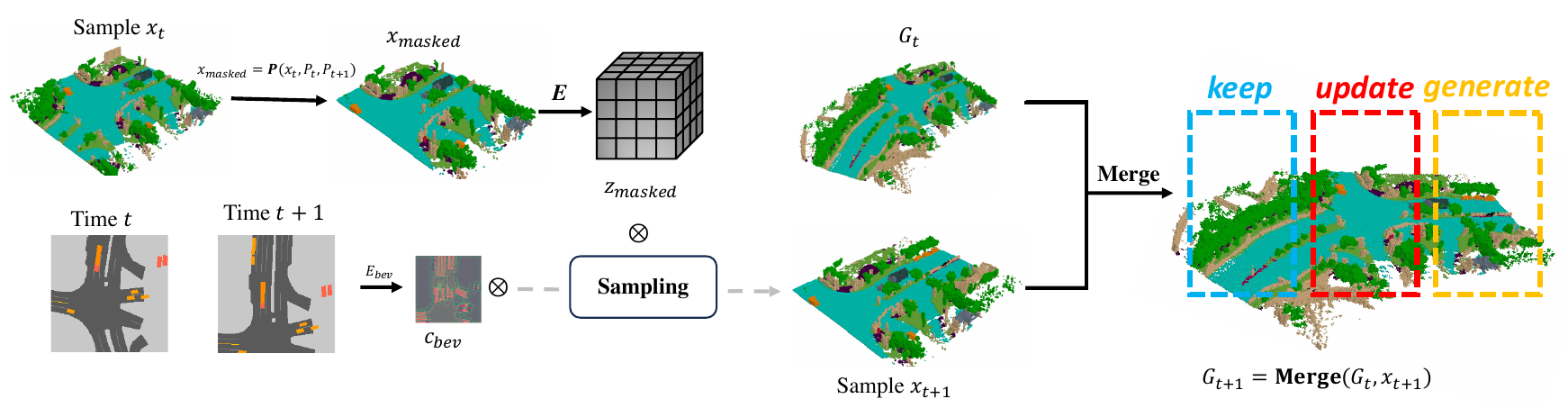}
    \caption{
    \textbf{Illustration of the scene expansion.} After projecting the generated sample $\mathbf{x}_t$ to the next frame via the ego poses $P_t$ and $P_{t+1}$ at time $t$ and $t+1$, and the BEV maps, we could get the overlap part and further encode both the $\mathbf{x}_{masked}$ and BEV to guide the generation process for the output sample $\mathbf{x}_{t+1}$ with high temporal consistency. Finally, we merge the sample $\mathbf{x}_{t+1}$ into the global scene $G_t$ with '\textit{keep}' for the original scene , \textit{'update'} for the intersection part by re-registering the labels of occupancy grids and \textit{'generate'} for the new part. }
    \label{fig:infinitegen}
    % \vspace{-3mm}
\end{figure}

It is intractable for a generative model to have a single-shot generation for a large-scale complicated 3D scene. We thus follow a divide-and-conquer strategy to tackle large-scale scene generation by designing a scene extension module. We first divide a large-scale scene into several single-frame parts and then separately generate the whole representation. After that, we have a scene extension step to extend and aggregate several single-frame scenes together. It is crucial to maintain the temporal consistency for the whole scene representation during the generation to ensure smooth expansion. The process of scene expansion is illustrated in Fig.~\ref{fig:infinitegen}. 
Given the observation $G_{t}$ of the scene at time $t$, we formulate the generation $\mathbf{x}_t$ at the time $t$ as:
\begin{equation}
    \mathbf{x}_{t+1} \sim P(\cdot|G_{t}),\  G_{t+1} = \textbf{Merge}(G_{t}, \mathbf{x}_{t+1}).
\end{equation}
The overlap between the current observation and the generated part is denoted as:
\begin{equation}
    \mathbf{x}_{masked} = \textbf{P}(\mathbf{x}_t,P_t,P_{t+1}),
\end{equation}
where $\textbf{P}$ means the projection through the current pose $P_{t+1}$ and the previous pose $P_{t}$.
One naive out-painting way is to directly utilize the generative capability of the diffusion model~\cite{repaint}. Given a fixed mask $M \in \{0,1\}^{h \times w \times d}$, the painting process becomes:
\begin{equation}
    \mathbf{z}_{t-1}=(1-M) \odot \mathbf{z}_{t-1}^{known}+M \odot \mathbf{z}_{t-1}^{unknown},
\end{equation}
where $\mathbf{z}_{t-1}^{known}$ is sampled from the distribution $\mathcal{N}(\sqrt{\bar{\alpha}_t}\mathbf{z}_0,(1-\bar{\alpha}_t) I)$ and $\mathbf{z}_{t-1}^{unknown}$ is sampled from distribution $\mathcal{N}(\mu_{\theta}(\mathbf{z}_t,t,c),
\Sigma_{\theta}(\mathbf{z}_t,t,c))$.  However, we found that during the progressive generation, it can accumulate errors and generate artifact results. So we take the mask image as another condition and fine-tune the pre-trained diffusion model $\theta$ to allow our model to generate a better result for scene expansion. The target distribution is formulated as:
\begin{equation}
    \mathrm{E}_{\mathbf{z}, t, \boldsymbol{\epsilon} \sim \mathcal{N}(\mathbf{0},\mathbf{I})}\|\boldsymbol{\epsilon} - \boldsymbol{\epsilon}_{\theta^{\prime}}(\mathbf{z}_t,t , c_{bev}, c_{mask}) \|.
\end{equation}
The progressive generation process thus can extend a single scene into a large-scale scene seamlessly in an unbounded way. %we could expand our scene to a large-scale representation in Fig.\ref{fig:infinitegen}. 

\subsection{Scene Image Synthesis}
The proposed 3D scene diffusion model not only generates diverse and realistic scenes based on the input condition but also can be used as a generative prior for many downstream applications. As follows, we describe how the learned model can be incorporated with pre-trained image generator for scene image generation. 

Recently, there are many attempts at generating 3D scenes conditioned on 3D voxels~\cite{infinicity,gancraft,scenedreamer}. These works have shown that explicit 3D structure can work effectively as a geometry prior to benefit 3D-aware scene image synthesis.
Following this line of works, we also showcase the utility of semantic occupancy grids generated by \NAME for scene image synthesis. These grids serve as an additional source of information, enabling improved synthesis of scene images.

To generate a scene image based on the generated semantic occupancy grid, our approach involves several steps. Firstly, we position a camera inside each scene. Next, we calculate the grids that intersect with the rays originating from the camera, and assign appropriate encodings to these grids. The encoding consists of two components: positional encoding of the grid's world coordinate, and semantic embedding. To determine the colors, we employ an MLP that decodes the encoding information.

To optimize our scene synthesis, we adopt the Score Distillation Sampling (SDS) technique, as introduced in DreamFusion~\cite{dreamfusion}. This involves distilling knowledge from a 2D diffusion model to refine our scene generation process. Additionally, we achieve controllable generation by fine-tuning a diffusion model that is conditioned on semantic and depth maps. To accomplish this, we gather a small set of driving images paired with perspective view images, semantic maps, and depth maps. Using this dataset, we fine-tune a pre-trained diffusion model~\cite{stablediff} to improve the quality and controllability of our scene synthesis. More details could be found in the supplementary.

\section{Experiments}
\noindent{\textbf{Dataset}} We evaluate our proposed method on the nuScenes Dataset~\cite{nuscenes} that contains multi-sensory data, High-precision maps, and 3D annotated labels. Two important data resources in our methods are 3D semantic occupancy grids with 2D synchronized BEV maps. We construct BEV maps following the instruction of ~\cite{cvt} by projecting the annotated objects onto the 2D HD maps and then transferring all the representations to the ego-frame coordinate system. To get 3D semantic occupancy grids for the whole scenes, we notice that there are some available datasets ~\cite{scene_as_occ,occ3d,surroundocc} aimed at occupancy prediction, which provides occupancy grids with annotated labels, achieved by accumulating all the collected LiDAR points, building meshes from these points and assigning semantic labels through calculating chamfer distance to the unlabeled points. The data of semantic occupancy map are collected from the dataset ~\cite{occ3d}, which allows us to train and evaluate our generative model. 

\noindent{\textbf{Implementation details}} At the training stage of the 3D VQVAE, we apply weighted cross entropy loss on the raw data and update the model weights in the mode of Exponential Moving Average. We crop the input data to the size of $192\times 192\times 16$ at the center and compress the data to a latent space with a dimension of $48\times 48 \times 4$. For the BEV map, an encoder is trained with unet to encode the BEV representation at the same resolution as the latent feature in the horizontal dimension. The diffusion model $\epsilon_\theta$ is then trained in the latent space for 100 epochs. Another 50 epochs are utilized to fine-tune the model $\epsilon_\theta$ to get model weights of completion model $\epsilon_\theta^{\prime}$.  We adopt DDIM~\cite{ddim} to facilitate the sampling stage with 100 steps. We use the $L_2$ norm for all distance calculations in noise prediction. Detailed ablation is shown in the supplementary.

\subsection{BEV-conditional Generation}
The BEV map, which describes a traffic situation at a particular location, is readily available in many driving simulators such as MetaDrive~\cite{metadrive} and Waymax~\cite{waymax}. With the designed model and pipeline, we get easily generate the corresponding urban scene with geometry and semantic information via the BEV map.
\begin{figure}[ht!]
\centering
\vspace{-3mm}

\includegraphics[width=1\linewidth,clip,trim = 0 0 0 0]{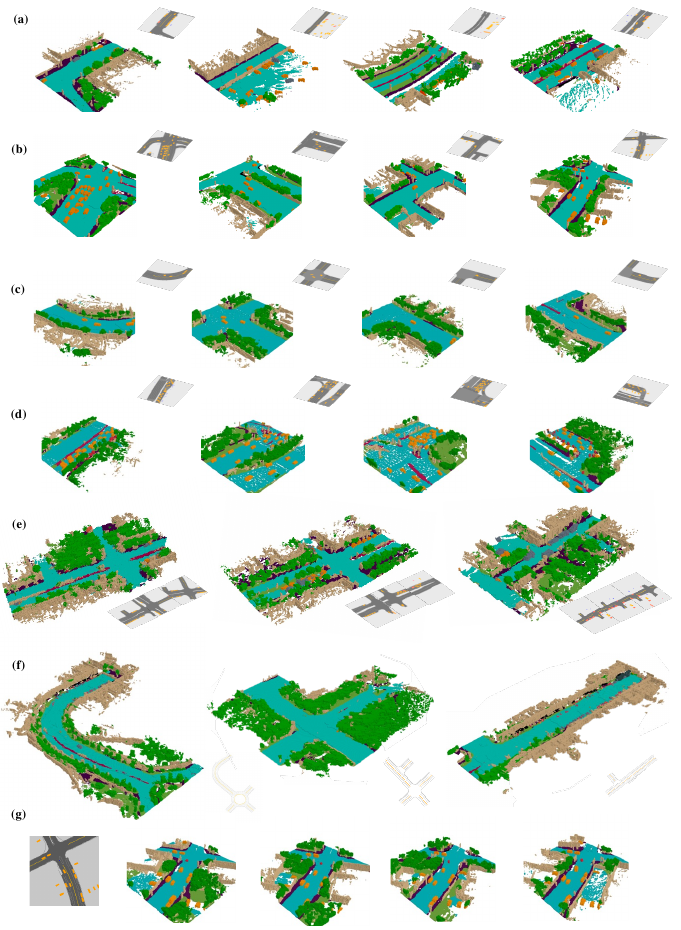}

\caption{Scenes generated from BEV maps sampled from nuScenes validation set (a), from Waymo Motion Dataset~\cite{waymo}(b), procedurally generated by MetaDrive simulator~\cite{metadrive} (c),  and from nuPlan Dataset~\cite{nuplan}. Large-scale scenes are generated from bev maps extracted from driving logs in nuScenes (e) and MetaDrvie simulator(f).  We also demonstrate diverse scenes generated conditioned on same BEV maps in (g).}
\label{fig:mainresult}
\vspace{-7mm}
\end{figure}

Fig.~\ref{fig:mainresult} shows several examples of the large-scale scenes generated based on the BEV maps randomly selected from the nuScenes validation set, the BEV maps synthesized from a driving simulator MetaDrive~\cite{metadrive} and the BEV maps from Waymo Motion Dataset~\cite{waymo}. By randomly sampling the model multiple times, we can also generate different diverse scenes from the same BEV map, as shown in Fig.~\ref{fig:mainresult}d.

It is interesting to see that even though without the data from MetaDrive~\cite{metadrive} for training, the trained model can generalize to the fake BEV map synthesized by the driving simulator and generate realistic scene accordingly. It demonstrates that the proposed \NAME is applicable to create new driving environments for interactive simulation. In particular, the generated scene already contains geometry and semantics, which is easy to incorporate in a driving simulator like MetaDrive.

\subsection{Quantitative Evaluation}
We build several evaluation metrics for our methods and baselines, since the task we propose to work on has not been widely explored before. The evaluation metrics consider the data quality, the diversity of the generated data, and the consistency to the input condition. 
\begin{itemize}
    \item \textbf{Quality} To measure the data distribution distance, like P-FID in ~\cite{pointe}, we individually train a specific network to get latent representation for all the scenes. Based on this, we could calculate a voxel-wise Fréchet Inception Distance, namely \textbf{V-FID} to measure the distribution distance between the generated data and the real data. We utilize an autoencoder network to compress the scene representation into a lower dimensional space and then calculate the distance of distribution in this compressed space. This metric reflects both geometry and semantics quality in the feature space. We also include \textbf{Maximum
    Mean Discrepancy (MMD)}, which is a non-parametric distance between two sets of samples. It compares the distance between two sets of samples by measuring the mean squared difference of the statistics of the two.
    \item \textbf{Accuracy} We calculate the Intersection over Union (IoU) to determine whether a voxel is occupied or not, and the mean Intersection over Union (mIoU) to handle different categories of data. The mIoU includes all the 16 categories from the dataset~\cite{occ3d}.

    \item \textbf{Condition Consistency} It is essential to assess if the conditioned generated samples are in alignment with the specified condition. We achieve this by extracting the height of the ground level from the generated voxel. The accuracy is then calculated relative to the given Bird's Eye View (BEV) map. This process enables us to determine the degree of alignment between the generated samples and the provided condition.
\end{itemize}
We've gathered a variety of related methods as baselines for comparison. Our Bird's Eye View (BEV)-conditioned model is compared with two 3D generative models: a diffusion model-based 3D U-Net, and a 2D-3D lifting method that separately generates a top-down view semantic image, a corresponding height image, and lifts these representations into 3D space. Such generation strategies have been previously explored either parametrically~\cite{infinicity} or non-parametrically~\cite{scenedreamer}. For each method, we generate 5k samples and evaluate these against data from the nuScenes validation set. Our evaluation incorporates the Variation of Fréchet Inception Distance (V-FID) to measure the distance of feature distribution, along with other metrics for scene-level evaluation. The human evaluation was conducted by asking 8 users to choose the better one when comparing visualization results. Each person evaluated 32 examples. The results demonstrated in Table~\ref{tab:baseline}, reveal that our proposed method perfors better in terms of both quantitative indicators (including V-FID and Maximum Mean Discrepancy (MMD)) and qualitative indicators (human evaluation).

\begin{table}[ht!]
    \centering
     \vspace{-3mm}
    \begin{minipage}[t]{0.48\textwidth}
    \centering
    \setlength{\tabcolsep}{12pt}
    
    \resizebox{1\linewidth}{!}{
    \begin{tabular}{lcc}
    \toprule
    Methods &   V-FID (\textdownarrow)  & MMD (\textdownarrow)    \\ 
    \toprule
    2D-3D lifting &567.0 &  1.292 \\
    % VQGAN ~\cite{taming} &   &&  \\ 
    Diff w. 3D U-Net ~\cite{stablediff} & 445.6  &  0.273 \\ 
    Ours & \textbf{291.4} & \textbf{0.106} \\
    \bottomrule
    \end{tabular}
    }
    % \caption{\textbf{a}}
    
    % \label{tab:baseline}

    \end{minipage}
    \begin{minipage}[t]{0.48\textwidth}

    \centering
    \setlength{\tabcolsep}{12pt}
    \resizebox{1\linewidth}{!}{
    \begin{tabular}{lc}
    \toprule
    Comparison & Preference \\
    \midrule
    Ours vs 2D-3D lifting & 100\% \\ 
    Ours vs Diff w. 3D U-Net & 96.9\% \\
    \bottomrule
    \end{tabular}
    }

    % \caption{\textbf{b}}

    \end{minipage}
    \vspace{+2mm}
    \caption{\textbf{Quantitative comparisons with different baselines}(left). V-FID and MMD are reported as evaluation metrics. and \textbf{Human Evaluation}(right).}
    \label{tab:baseline}
    \vspace{-12mm}
\end{table}

\begin{figure}[ht!]
    \centering
    \includegraphics[width=1\linewidth]{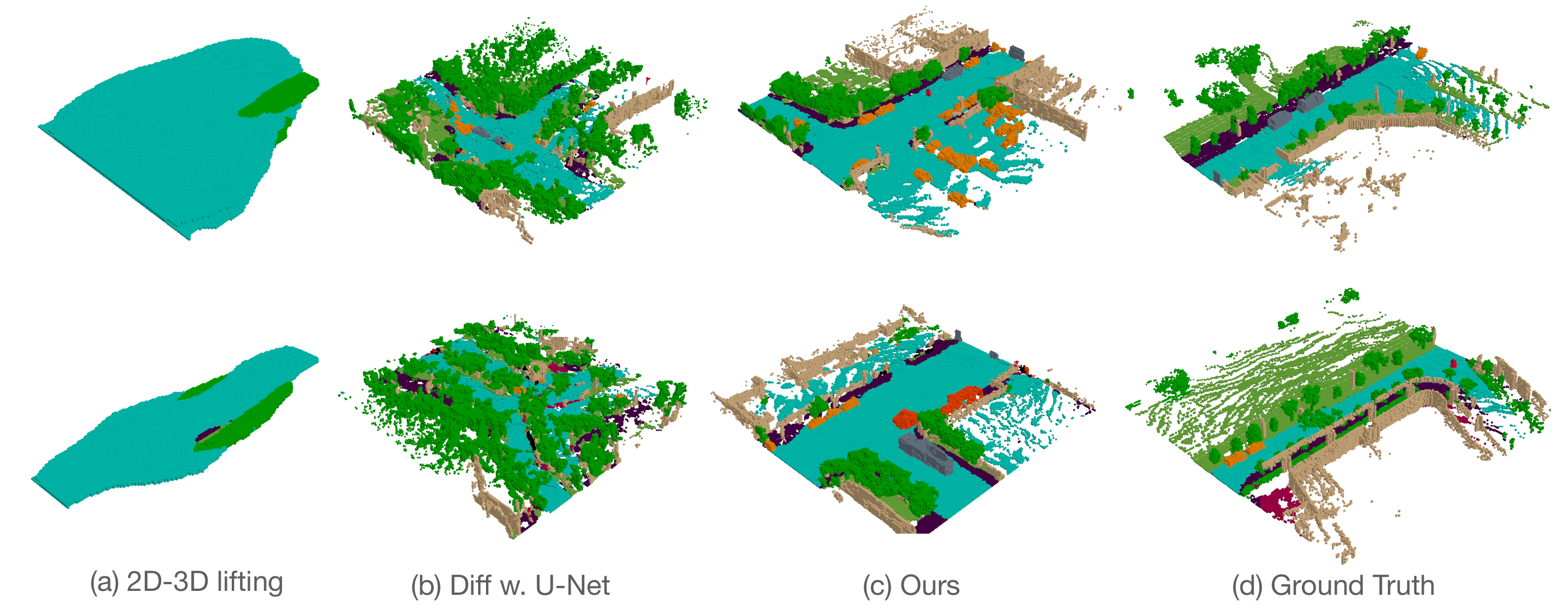}
    \caption{Qualitative examples for different methods and ground truth data.}
    \label{fig:vis_comp}
\end{figure}

\subsection{Ablation study}
\noindent{\textbf{Autoencoder}}
The selection of the initial stage model is a critical factor. For this reason, we have conducted an ablation study on different architectural designs for the Vector Quantized Variational AutoEncoder (VQ-VAE) model. As demonstrated in Table~\ref{tab:ablation_ae}, certain types of VQ-VAE show superior performance compared to those applying a KL divergence penalty, in terms of Intersection over Union (IoU) and mean IoU (mIoU). This suggests that the discrete variant of VQ-VAE is better suited for 3D occupancy data. Additionally, an increase in the embedding dimension results in an enhancement of IoU and mIoU. Downsampling to 1/8 significantly impairs performance. Therefore, based on the findings outlined in Table~\ref{tab:ablation_ae}, we opt to downsample the raw data to a 1/4 resolution and use a codebook with 2048 embeddings.

\begin{table}[ht!]
    \centering
    \begin{minipage}[t]{0.50\textwidth}
    \setlength{\tabcolsep}{12pt}
    % \renewcommand\arraystretch{1.2}
    % \footnotesize
    \resizebox{1\linewidth}{!}{
    \begin{tabular}{lcc}
    \toprule
    Methods & IoU (\textuparrow)  & mIoU (\textuparrow)  \\ 
    \toprule
    VAE, KL & 0.982 & 0.682 \\
    VQVAE, embed 512 &  0.987 & 0.764 \\ 
    % VQVAE, embed 1024 (without ema + weighted loss)& 0.986 & 0.784 \\ 
    VQVAE, embed 1024 with 1/4 resolution & \textbf{0.988} & 0.798 \\ 
    VQVAE, embed 2048 with 1/8 resolution &  0.973 &  0.513 \\
    VQVAE, embed 2048 with 1/4 resolution & \textbf{0.988} & \textbf{0.800} \\ 
    \bottomrule
    \end{tabular}
    }
    \vspace{2mm}
    \caption{\textbf{Comparisons with different autoencoders. }IoU and mIoU are utlized to evluate the reconstruction results. }
    \label{tab:ablation_ae}
    \end{minipage}
    \centering
    \begin{minipage}[t]{0.45\textwidth}

    \centering
    \setlength{\tabcolsep}{12pt}
    \resizebox{1\linewidth}{!}{
    \begin{tabular}{lcc}
    \toprule
    Methods & V-FID (\textdownarrow) & CC (\textuparrow) \\ 
    \toprule
    Crossattn & 311.3 &  0.252\\
    Modulation &317.4 &  0.674\\
    Concat &  \textbf{291.4 } & \textbf{0.789}\\ 
    \bottomrule
    \end{tabular}
    }
    \vspace{2mm}
    \caption{\textbf{Comparisons with different condition options.} V-FID and CC (condition consistency) are reported as evaluation metrics.} 
    \label{tab:ablation_cond}
    \end{minipage}
    \vspace{-7mm}
\end{table}

\noindent{\textbf{Different conditioning options}} In our study, we conduct an experiment to determine the most effective method for integrating the Bird's Eye View (BEV) feature into the diffusion process. We consider three potential options: modulation, cross-attention, and concatenation. The modulation approach is similar to Adaptive Instance Normalization (AdaIN)\cite{adain}, the cross-attention method borrows from the text-to-image synthesis technique\cite{stablediff}, and the concatenation strategy aligns the 2D BEV feature with the 3D latent space by pooling along the height axis.

\begin{figure}[ht!]
\centering

\includegraphics[width = 1\linewidth]{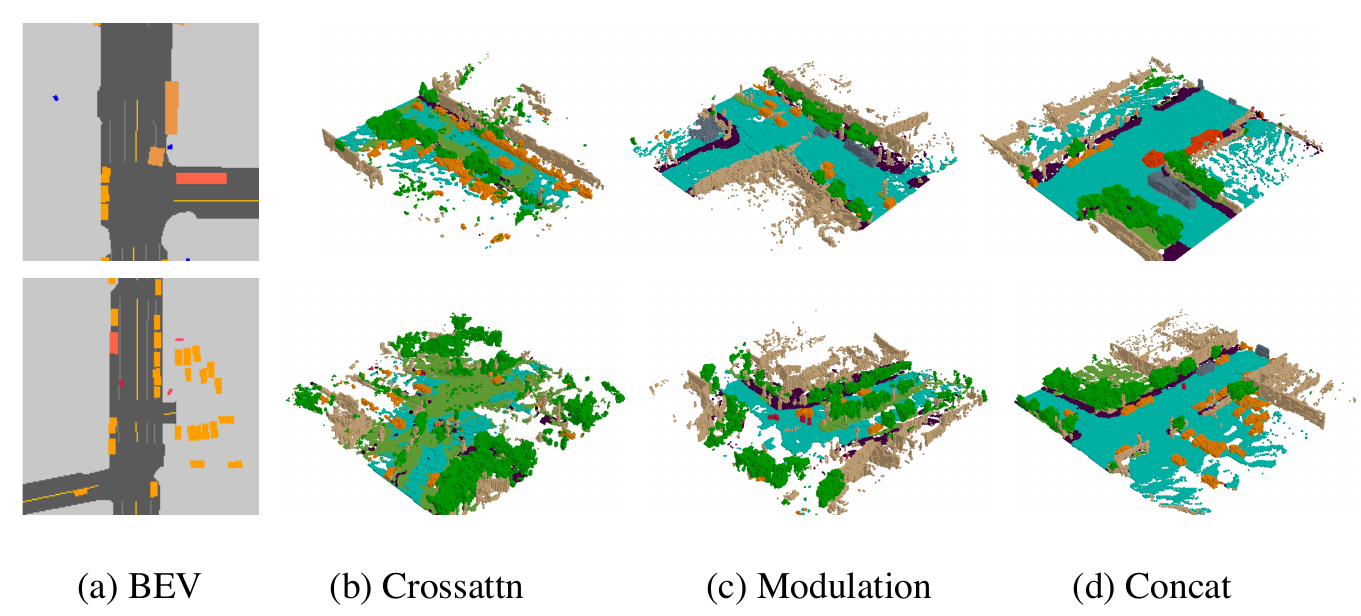}
\caption{Comparison of different methods of incorporating conditions. Compared to the cross-attention and adaIn methods, concatenating the BEV (Bird's Eye View) condition can achieve superior controllability during the generation process.}
\label{fig:bev_cond}
\vspace{-12mm}
\end{figure}

To measure the efficacy of these conditional generation strategies, we measure the V-FID scores, which serve as an indicator of both the quality of the generated samples and the consistency with the given condition. As shown in Fig.\ref{fig:bev_cond} and Table~\ref{tab:ablation_cond}, the concatenation strategy, which combines the BEV feature with the latent feature, outperforms both the cross-attention and modulation methods. This finding implies that concatenation more effectively aligns the 2D BEV features with the 3D representation, thereby ensuring a more accurate and reliable transformation from the 2D BEV map to the 3D generated output. Consequently, our results suggest that the concatenation strategy is an optimal choice for integrating BEV features into the diffusion process when generating 3D data from 2D conditions.

\subsection{Point cloud segmentation}
We take the experiment of point cloud segmentation as a downstream task for our generation to verify the effectiveness of our generated data. We convert all the occupancy grids into point clouds, and then consider 3D semantic segmentation and verify the augmentation using our synthetic data via Cylinder3D~\cite{cylinder3d}. Tab.~\ref{tab:pcdseg} shows generated data could boost the performance of the perception model and verifies the effectiveness of our data.
\begin{table}[ht!]
\scriptsize

\vspace{-2mm}
\setlength{\tabcolsep}{0.005\linewidth}
\newcommand{\classfreq}[1]{{~\tiny(\semkitfreq{#1}\%)}}  

\def\mystrut{\rule{0pt}{1.5\normalbaselineskip}}
\centering
\resizebox{\textwidth}{!}{

\begin{tabular}{l |  c c c c c c c c c c c c c c c c |c}

    \toprule
    Method 
    % & \rotatebox{90}{others} 
    & \rotatebox{90}{barrier}
    & \rotatebox{90}{bicycle} 
    & \rotatebox{90}{bus} 
    & \rotatebox{90}{car} 
    & \rotatebox{90}{Cons. Veh} 
    & \rotatebox{90}{motorcycle} 
    & \rotatebox{90}{pedestrian} 
    & \rotatebox{90}{traffic cone} 
    & \rotatebox{90}{trailer} 
    & \rotatebox{90}{truck} 
    & \rotatebox{90}{Dri. Sur} 
    & \rotatebox{90}{other flat} 
    & \rotatebox{90}{sidewalk} 
    & \rotatebox{90}{terrain} 
    & \rotatebox{90}{manmade} 
    & \rotatebox{90}{vegetation} 
    & \rotatebox{90}{mIoU}  \\
    \midrule
% barrier : 42.57%
% bicycle : 7.70%
% bus : 58.04%
% car : 79.40%
% construction_vehicle : 27.95%
% motorcycle : 30.46%
% pedestrian : 52.11%
% traffic_cone : 11.12%
% trailer : 36.87%
% truck : 40.47%
% driveable_surface : 81.47%
% other_flat : 29.57%
% sidewalk : 46.17%
% terrain : 52.45%
% manmade : 83.59%
% vegetation : 88.09%
% Current val miou is 48.002 while the best val miou is 48.002
    Real &  42.57   & 7.70 & 58.04 & 79.40 & 27.95 & 30.46 & 52.11 & 11.12 & 36.87 & 40.47 & 81.47 & 29.57 & 46.17 & 52.45 & 83.59 &  88.09 & 48.00 \\
% barrier : 43.79%
% bicycle : 7.32%
% bus : 70.82%
% car : 79.15%
% construction_vehicle : 29.48%
% motorcycle : 30.67%
% pedestrian : 53.32%
% traffic_cone : 10.21%
% trailer : 41.49%
% truck : 53.29%
% driveable_surface : 82.47%
% other_flat : 32.13%
% sidewalk : 47.18%
% terrain : 54.22%
% manmade : 83.81%
% vegetation : 88.35%
% Current val miou is 50.482 while the best val miou is 50.482
    Real w. aug.  & \textbf{43.79} & 7.32  & \textbf{70.82}  & 79.15  & \textbf{29.48}  & 30.67  & \textbf{53.32}  & 10.21  & \textbf{41.49 } & \textbf{53.29}  & \textbf{82.47} &\textbf{ 32.13} & \textbf{47.18} & \textbf{54.22} & 83.81 & 88.35& \textbf{50.48} \\

\bottomrule
\end{tabular}
}
\vspace{2mm}
\caption{3D point cloud segmentation experiment. We show the IoU results for all the categories on the nuScenes validation set. The segmentation experiment verfy the effectiveness of our generated data which could boost the performance of the downstream task. Those results that differ by more than one percentage point are marked.}
\label{tab:pcdseg}
\vspace{-14mm}
\end{table}
\subsection{Scene Image Synthesis}
\label{sec:scene_synth}
\begin{figure}[ht!]
    \centering
    \includegraphics[width=1\linewidth]{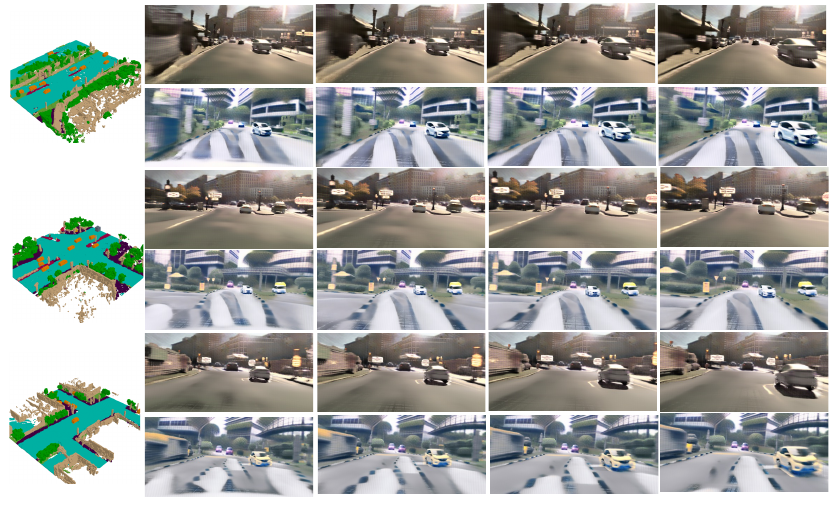}
    \caption{\textbf{Scene image synthesis based on semantic occupancy grid.} The left column illustrates the generated semantic occupancy grid observed in overhead view. The right columns show the synthesized scene images from different angles. The text descriptions for each scene are 'a driving scene image at Boston, daytime.' and 'a driving scene image at Singapore, daytime.'}
    \label{fig:image_syn}
    \vspace{-10mm}
\end{figure}

We further verify that the generated semantic occupancy grids could be used as a generative prior and benefits the scene image synthesis. 
Specifically, we use Score Distillation Sampling to distill knowledge from a 2D diffusion model to optimize the rendered urban scene.
By parameterizing the scene using an MLP as the renderer and optimizing it using SDS, we are able to transform the semantic occupancy grid into a 3D environment with visual appearance. The result is shown in Fig.~\ref{fig:image_syn}. We can see that the synthesized images exhibit visual diversity and content consistency with the semantic labels.

\section{Conclusion}

We present \NAME, a 3D diffusion approach for generating unbounded urban scenes. The proposed model accurately represents the geometry and semantics of objects by conditioning on BEV layout, producing diverse and realistic urban scenes in the form of semantic occupancy maps. The model also demonstrates the capability to generate realistic urban scenes based on the BEV layouts synthesized from a driving simulator. 
\paragraph{Limitations.} The visual appearance of the synthesized scene image is not ideal, as the image synthesis is not the focus of this work and is independent of the 3D diffusion model training. In the future, we will train a multi-modal diffusion model that can generate geometry, semantics, and visual appearance at the same time. On the other hand, there is no object instance in the generated scene and the moving objects in the scene also bring some artifacts. We will incorporate instance information in the diffusion process in the future.

\bibliographystyle{splncs04}
\bibliography{main}

% \newpage
% \appendix
% \begin{center}
% \bf{\LARGE APPENDIX}
% \end{center}
% \input{files/supp}
\end{document}